\documentclass{article}
\usepackage{spconf,amsmath,graphicx}
\usepackage{pstricks}
\usepackage{url}

\title{Residual and Plain Convolutional Neural Networks for 3D Brain MRI Classification}
\name{Sergey Korolev$^{\star \dagger}$ \qquad Amir Safiullin$^{\dagger}$ \qquad Mikhail Belyaev$^{\star \dagger}$ \qquad Yulia Dodonova$^{\dagger}$}

\address{$^{\star}$ Skolkovo Institute of Science and Technology \\
    $^{\dagger}$ Institute for Information Transmission Problems \\
    {\footnotesize \texttt{s.korolev@skoltech.ru, aesafiullin@edu.hse.ru, m.belyaev@skoltech.ru, dodonova@iitp.ru}}}
\begin{document}
%\ninept
%
\maketitle
% rewrote abstract a bit
\begin{abstract}
In the recent years there have been a number of studies that applied deep learning algorithms to neuroimaging data. Pipelines used in those studies mostly require multiple processing steps for feature extraction, although modern advancements in deep learning for image classification can provide a powerful framework for automatic feature generation and more straightforward analysis. In this paper, we show how similar performance can be achieved skipping these feature extraction steps with the residual and plain 3D convolutional neural network architectures. We demonstrate the performance of the proposed approach for classification of Alzheimer's disease versus mild cognitive impairment and normal controls on the Alzheimer’s Disease National Initiative (ADNI) dataset of 3D structural MRI brain scans.
\end{abstract}

\section{Introduction}
\label{sec:intro}

A growing number of machine learning studies based on neuroimaging data aim to  both develop diagnostic tools that help brain MRI classification and automatic volume segmentation, and understand the mechanics of diseases, including the neurodegenerative ones. Recently, there appeared several publications that use deep learning for 3D structural magnetic resonance images (MRI) processing, with particular algorithms ranging from manifold learning with Restricted Boltzmann Machines~\cite{Brosch2013} to a combination of autoencoders and classifiers~\cite{Liu2014} and to meta analysis of different state of the art approaches~\cite{Plis2014}.

% removed "rare", fix here for handcrafted feature generation + extraction
In these studies, a problem of MRI classification is usually tackled with complex multistep pipelines for handcrafted feature generation and feature extraction from the data that precede standard machine learning techniques such as support vector machines (SVM) or logistic regression used for classification. In our study, we develop deep learning based algorithms that have a potential to overcome this problem with end-to-end models and simplify the MRI classification pipeline.

Also, it is important to note that datasets collected in neuroimaging studies are commonly very small, compared to the sizes of image classification datasets that are currently used to train neural networks for object classification and detection in 2D image analysis. Although there are recent studies aiming to overcome this issue by means of oversampling (e.g., ~\cite{Plis}),  in supervised learning setup it is still incredibly important to be able to build network architectures capable to learn features necessary for classification based on small datasets.

In this paper, we propose two different 3D convolutional network architectures for brain MRI classification, which are the modifications of a plain and residual convolutional neural networks.  We choose to use convolutional neural networks for their ability to tackle the two problems stated above. First, these networks can generalize local features into a meta-representation of an object for image recognition or classification. Second, modern advancements in deep learning for image classification such as batch normalization technique and residual network architectures relieve the issues of having small training datasets, while providing a powerful framework for automatic feature generation. As a result, we have models that can be applied to 3D MRI images without intermediate handcrafted feature extraction.

We examine the performance of the proposed network architectures based on the data stemming from the Alzheimer's Disease Neuroimaging Initiative (ADNI) project that provides a  dataset of structural MRI scans with labels and subject metadata. We use this dataset to test our models' performance for a task of classifying MRI scans of subjects with Alzheimer’s disease (AD), early and late mild cognitive impairment (EMCI and LMCI), and normal cohort (NC).

There are previous studies that used deep learning algorithms for  Alzheimer's Disease classification. For example, Heung-Il Suk et al.~\cite{Suk2014} used a complex architecture of deep belief network (DBN) with patch-sampling preprocessing and a combination of MRI and PET modalities with a weighted ensemble of SVM classifiers on top of the network. Preprocessing included skull stripping and cerebrum removal, segmentation into grey matter, white matter and cerebrospinal fluid (CSF), and sampling patches as neighborhoods of statistically significant voxels. The resulting classification performance for AD/NC, MCI/NC and MCI-C/MCI-NC was .95, .85, .76 accuracy and .99, .88, .75 ROC AUC.

In a recent paper on ADNI data classification, Ehsan Hosseini-Asl et al.~\cite{Hosseini-Asl2016a} proposed to use a 3D convolutional neural network for feature extraction from MRIs.  To be more specific, the authors of ~\cite{Hosseini-Asl2016a} used the Deeply Supervised Adaptive 3D-CNN (DSA-3D-CNN) which was initialized by training convolutional autoencoders for feature extraction and fine tuning the network for classification on different domain images. They show impressive performance compared to the other approaches with binary ROC AUC over $.96$. This approach is the closest to the one we propose, but it has a few caveats. Although  for some types of data this approach may yield better initial performance, it has downsides in the time complexity of weight initialization.

An important note is that previous ADNI-based studies report results of a binary classification for the pairs of available classes, not a multiclass classification. In our study, we also test the performance of the proposed models in a task of binary one-versus-one classification  so as to be able to compare our results with previous works. We demonstrate that our approach achieves relatively good results without complex preprocessing or model stacking while showing a range of possible avenues for further research both in terms of data augmentation and oversampling and in terms of model architecture design.

% Eduardo Castro et al.~\cite{Plis} describe an oversampling technique for MRI classification. They start with grey matter extraction and then unravel the intensities of voxels into the 60465-dimensional vector. Then they apply independent component analysis technique to factorize the data matrix into two, with one corresponding to subjects and the other -- to voxels. After that, they estimate the pdf of each normalized latent variable of first matrix across the given subject class and sample synthetic value from this distribution. In the end, they multiply this vector by the second matrix to get the synthetic sample. They evaluate the performance of this oversampling technique by measuring F1 score of binary classification task for KNN, linear SVM and logistic regression.

\section{Method}
\label{sec:method}

We compare two different approaches to brain MRI classification with convolutional neural networks. The first one is a common feedforward network with convolutional and pooling layers, and the second one is a modern residual neural network. For each architecture we run a number of experiments to determine the number of layers, number of filters for each layer, learning rate, pooling size, and dropout probability.

\subsection{VoxCNN}
We choose to use network architecture similar to that of VGG~\cite{Simonyan14c} for image classification to examine whether this type of convolutional network is able to extract features necessary for 3D image classification.

Our VoxCNN architecture has four volumetric convolutional blocks for extracting features (with a number of filters increasing from layer to layer), two deconvolutional layers with batchnorm and dropout for regularization and an output with softmax nonlinearity for classification (see Fig. \ref{fig:resnet} (a)). We train the final binary classification models using AdaM with learning rate of $27 * 10^{-6}$ and batchsize of $5$ for $150$ epochs.

\subsection{ResNet}
\label{ssec:resarch}

Residual neural networks~\cite{He2015} architecture won the Imagenet contest in 2015 and demonstrated possibility to greatly improve the depth of the network while having fast convergence. There are already publications that show their results for 3D image segmentation like VoxResNet~\cite{VoxResNet16}, where the authors use ideas of identity connections for ResNets~\cite{He2016}.

\begin{figure}[!htb]

\begin{minipage}[b]{0.48\linewidth}
  \centering
  \centerline{\includegraphics[width=2.67cm]{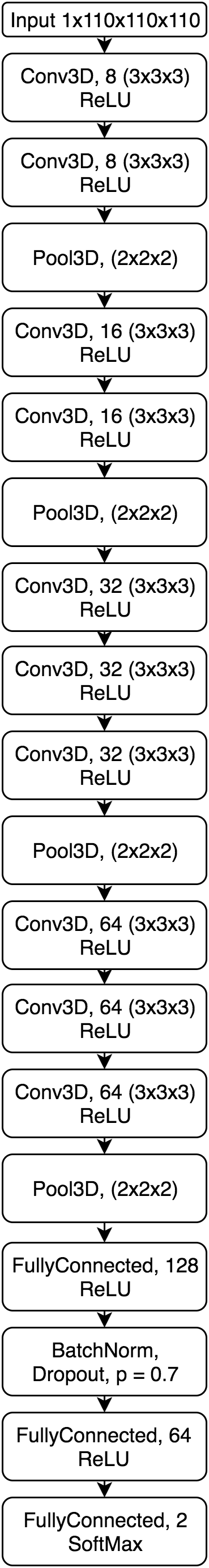}}
%  \vspace{1.5cm}
  \centerline{(a) VoxCNN}\medskip
\end{minipage}
\hfill
\begin{minipage}[b]{0.48\linewidth}
  \centering
  \centerline{\includegraphics[width=2.67cm]{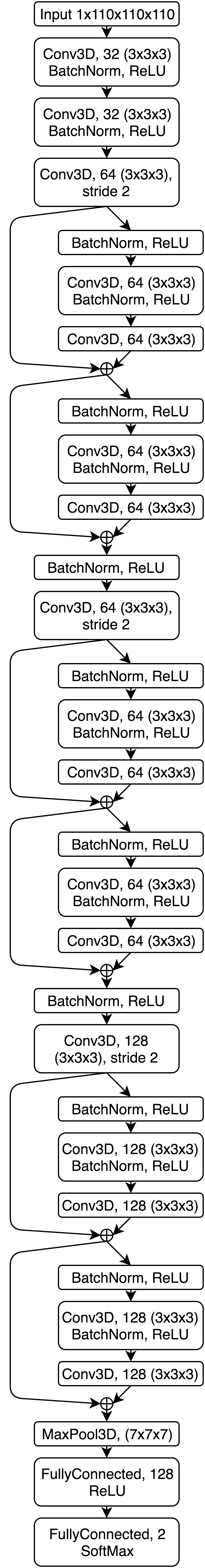}}
%  \vspace{1.5cm}
  \centerline{(b) ResNet}\medskip
\end{minipage}

\caption{VoxCNN and ResNet architectures.}
\label{fig:resnet}
\end{figure}

Our ResNet architecture derived from VoxResNet has 21 layers containing six VoxRes blocks each with 64 filters for convolution, except the last two, which has 128 and convolutions with $2 \times 2 \times 2$ strides between VoxRes blocks to reduce the dimension of the layer output. The output of the last VoxRes block is sent to a pooling layer to further reduce it to $2 \times 2 \times 2 \times 128$, followed by a fully connected layer with 128 hidden units and an output for binary classification with softmax nonlinearity (see Fig. \ref{fig:resnet} (b)). We train the final binary classification models using Nesterov momentum with learning rate of $10^{-4}$ and batchsize of $3$ for 70 epochs.

\subsection{Setup}
\label{ssec:setup}

We run 5-fold cross-validations with 5 different fold splits each time to get better approximation of prediction performance. On each fold we train the network for a fixed number of epochs for each classification task to get the perfect class separation on a training subset of the dataset and stabilize the performance metrics on validation subset.

It is also important to note that because of the model size and the limitations of GPU memory, we have to modify the batch iteration process so  that we are sure that there are samples of each class in every batch. The reason being that the probability of having only one class represented inside a batch for infinite number of samples is $\frac{1}{c^b}$ where $c$ is the number of classes and $b$ is the batchsize. Therefore, for large batchsizes this probability is low, but in our case it is high enough to destabilize the learning process. The balancing of the samples inside each batch gives us more stable learning curves.

\section{Data}
\label{sec:experiment}

Data used in preparation of this article were obtained from the Alzheimer's Disease Neuroimaging Initiative (ADNI) database (adni.loni.usc.edu). As such, the investigators within the ADNI contributed to the design and implementation of ADNI and/or provided data but did not participate in analysis or writing of this report. A complete listing of ADNI investigators can be found at \url{http://adni.loni.usc.edu/wp-content/uploads/how_to_apply/ADNI_Acknowledgement_List.pdf}

For our experiments we use subset of ADNI structural MRI data that has been preprocessed with alignment and skull-stripping marked as ``Spatially Normalized, Masked and N3 corrected T1 images''. Since there are patients that have multiple images taken during a period of time and we want to prevent possible information ``leaks'', we only select the first images taken for each subject. Resulting dataset has 231 images of four classes: 50 of Alzheimer's Disease (AD) patients, 43 of Late Mild Cognitive Impairment (LMCI), 77 of Early Mild Cognitive Impairment (EMCI) and 61 of Normal Cohort (NC). With these four classes, we have six binary (one-versus-one) classification tasks. All of the images are stored as voxel intensity values in 3D tensor of shape $110 \times 110 \times 110$.

\section{Results}
\label{sec:results}

The results for six binary classification tasks are shown in the Table~\ref{tab:res}. The network learns to accurately classify Alzheimer's Disease subjects from Normal Cohort, however struggles to separate them from intermediate classes of Late and Early Mild Cognitive Impairment. Both networks show similar results within a standard deviation.

\begin{table}[!h]
\footnotesize
\begin{tabular}{l|cc|cc}
  & \multicolumn{2}{c}{VoxCNN} & \multicolumn{2}{c}{ResNet} \\ \hline
   & AUC & Acc. & AUC & Acc. \\ \hline
  AD vs NC & $.88 \pm .08$ & $.79 \pm .08$ & $.87 \pm .07$ & $.80 \pm .07$ \\
  AD vs EMCI & $.66 \pm .11$ & $.64 \pm .07$ & $.67 \pm .13$ & $.63 \pm .09$ \\
  AD vs LMCI & $.61 \pm .12$ & $.62 \pm .08$ & $.62 \pm .15$ & $.59 \pm .11$ \\
  LMCI vs NC & $.67 \pm .13$ & $.63 \pm .10$ & $.65 \pm .11$ & $.61 \pm .10$ \\
  LMCI vs EMCI & $.47 \pm .09$ & $.56 \pm .11$ & $.52 \pm .11$ & $.52 \pm .09$ \\
  EMCI vs NC & $.57 \pm .12$ & $.54 \pm .09$ & $.58 \pm .09$ & $.56 \pm .07$ \\
\end{tabular}
\caption{Classification ROC AUC and accuracy (mean $\pm$ std.)}
\label{tab:res}
\end{table}
\normalsize

\begin{figure}[!htb]

  \centering
  \centerline{\includegraphics[width=8.5cm]{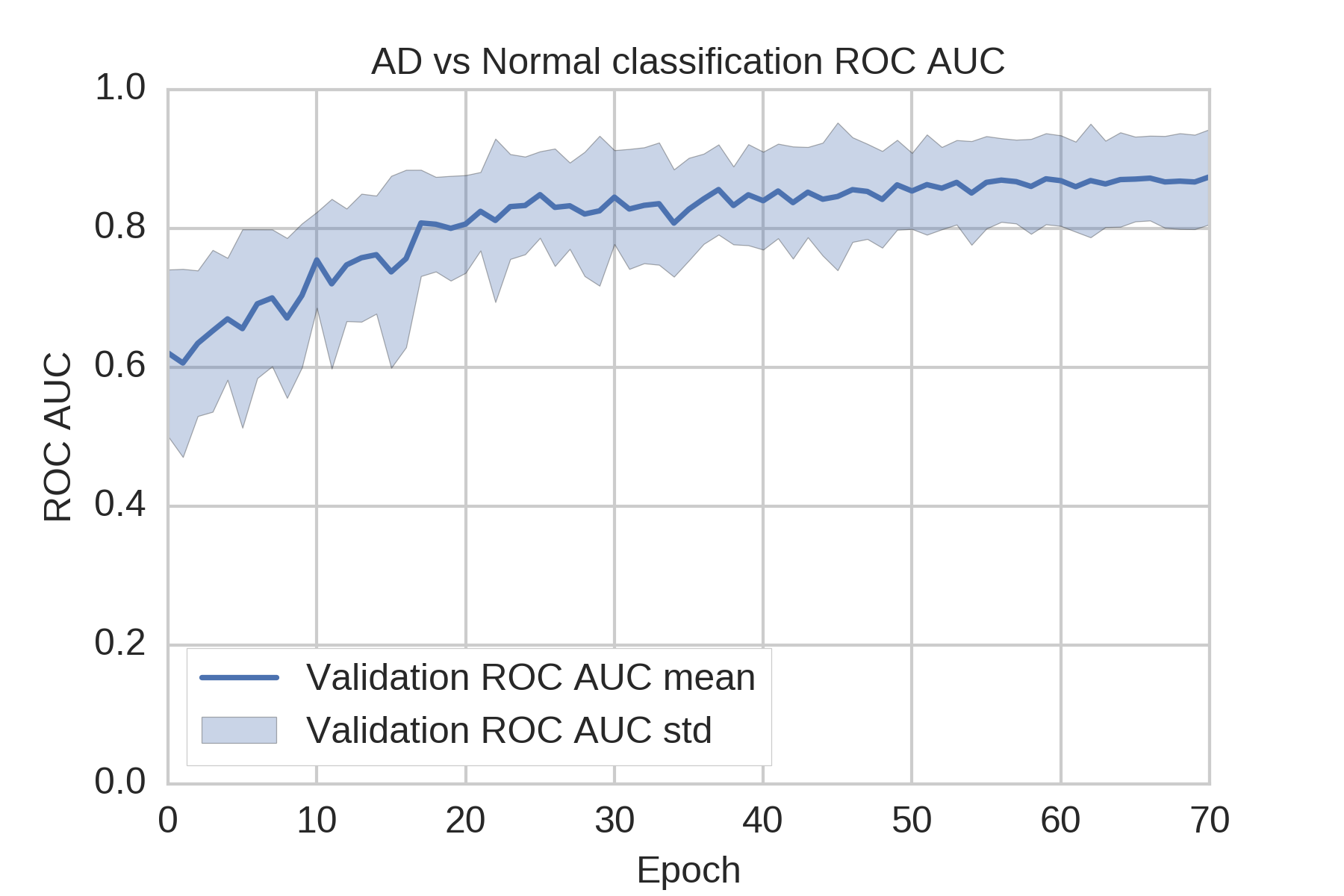}}

\caption{ROC AUC plot for AD vs Normal classification of ResNet.}
\label{fig:aucs}
\end{figure}

Figure~\ref{fig:aucs} shows the ROC AUC values for the validation dataset during training. This plot demonstrates that the network classification performance constantly improves and then plateaues after the 50th epoch.

\begin{figure}[!htb]

  \centering
  \centerline{\includegraphics[width=8.5cm]{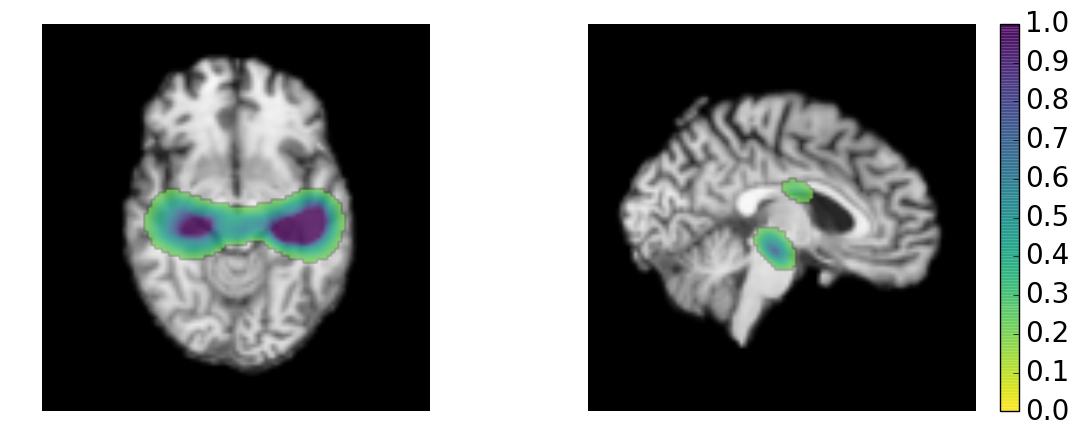}}

\caption{Network attention areas for Normal Cohort MRI. (Axial and sagittal view)}
\label{fig:attention}
\end{figure}

Figure~\ref{fig:attention} shows an example of network attention generated by prediction with obstructed images~\cite{Fergus13} for a Normal Cohort subject from a test subset. We produce it by compiling the prediction mask while obstructing image parts with $7 \times 7 \times 7$ box and measuring the drop of the output probability. This heatmap shows how the network learns the importance of the areas. These areas are blurry, but the spots with the highest attention seem to match with the ones most affected by the Alzheimer's Disease, mainly hippocampus~\cite{hippocamp11} and ventricles~\cite{vent10}.

\section{Conclusion}
\label{sec:conclusion}

We proposed deep 3D convolutional neural network architectures for a task of classification of brain MRI scans. We demonstrated performance of the residual and plain convolutional neural networks based on the ADNI dataset which is a largest available dataset of structural MRIs of subjects with Alzheimer’s disease and normal controls. We showed that applying s the proposed models to MRI classification problem yields results comparable to previously used approaches. The major advantages of our method are the ease of use and no need for handcrafted feature generation. The proposed approach should prove useful for on the fly prediction of any given MRI scan as soon as we can automatically process incoming images for skull stripping and normalization.

In our future work, we hope to achieve similar or better results for images that were not preprocessed for alignment and skullstripping, since convolutional neural networks are invariant to translation of the object on the image if they have global pooling after convolutions. This would mean a possibility  of one-step analysis of complex MRI data instead of multistep pipelines that currently dominate the field.

\bibliographystyle{IEEEbib}
\bibliography{strings,refs,bibliography}

\end{document}